\documentclass[twocolumn, 10pt]{article}

\usepackage{graphicx} 
\usepackage{etoolbox}
\usepackage{caption}
\usepackage{cite}
\usepackage{amsfonts, amsmath}
\usepackage{hyperref}  
\usepackage{geometry}
\usepackage{authblk}

\relax
\newcommand{\insertfig}{\includegraphics[width=\textwidth]{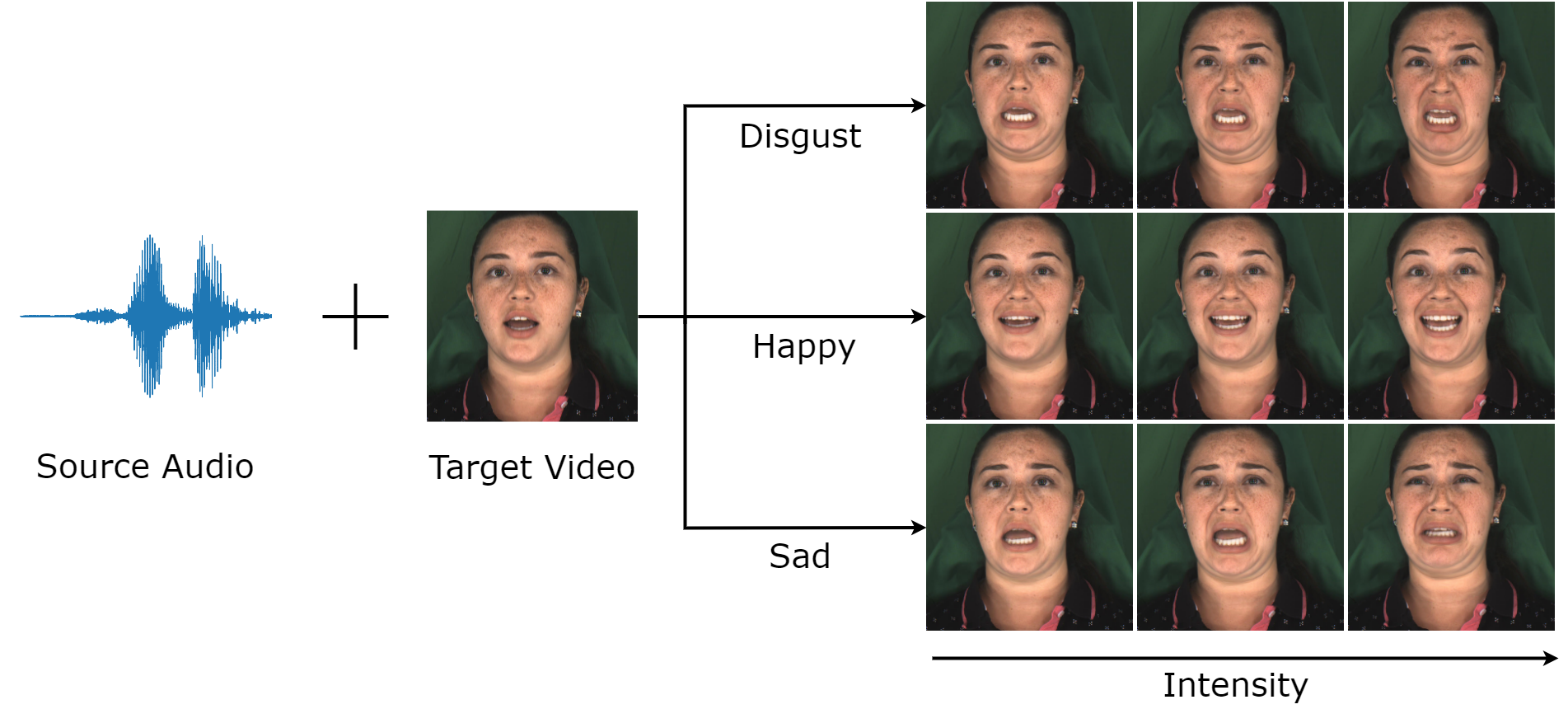}\captionof*{figure}{Our method, \textbf{READ Avatars} allows for the creation of photo-realistic and lip synchronized video from audio and a reference video, with control over emotion. The same audio can be used to generate videos in multiple emotions. The intensity of each emotion can be directly specified, allowing for fine-grained control over the output.}\vspace{5mm}}

\makeatletter
\apptocmd{\@maketitle}{\centering\insertfig}{}{}
\makeatother

\begin{document}

\title{READ Avatars: Realistic Emotion-controllable Audio Driven Avatars}
\date{}

\author{Jack Saunders*}
\author{Vinay Namboodiri}
\affil{Dept. of Computer Science, University of Bath, \texttt{\{jrs68, vpn22\}@bath.ac.uk}}

\maketitle

\begin{abstract}
\begin{quote}
We present READ Avatars, a 3D-based approach for generating 2D avatars that are driven by audio input with direct and granular control over the emotion. Previous methods are unable to achieve realistic animation due to the many-to-many nature of audio to expression mappings. We alleviate this issue by introducing an adversarial loss in the audio-to-expression generation process. This removes the smoothing effect of regression-based models and helps to improve the realism and expressiveness of the generated avatars. We note furthermore, that audio should be directly utilized when generating mouth interiors and that other 3D-based methods do not attempt this. We address this with audio-conditioned neural textures, which are resolution-independent. To evaluate the performance of our method, we perform quantitative and qualitative experiments, including a user study. We also propose a new metric for comparing how well an actor's emotion is reconstructed in the generated avatar. Our results show that our approach outperforms state of the art audio-driven avatar generation methods across several metrics. \footnote{A demo video can be found at \url{https://youtu.be/QSyMl3vV0pA}}
\end{quote}
\end{abstract}

\section{Introduction}

Generating convincing talking head video is a highly desired capability in various applications, such as film and television dubbing, video games and photo-realistic video assistants. While significant progress has been made in this area \cite{vougioukas2018end, DVP, NSPVD, Wav2Lip, ATVG, Sun2022, thies2020nvp, EAMM, Siarohin_2019_NeurIPS, guo2021adnerf, LiveSpeechPortraits, wen2020photorealistic, Tang2022, TextBasedEdit}, most existing methods produce either low-quality but accurate lip sync using 2D models \cite{Wav2Lip, ATVG, Sun2022, vougioukas2018end} or high-quality but inconsistent lip sync using 3D models \cite{DVP, NSPVD, thies2020nvp, LiveSpeechPortraits, wen2020photorealistic, Tang2022}. We hypothesize that two key factors have prevented the development of models that are both high-quality and lip synchronized. The first is that audio to expression is a many-to-many mapping. A given audio can correspond to many lip shapes, and the same lip shapes can produce different audio due to factors such as the larynx. The second factor is that while 3D models improve the visual quality by introducing strong priors, they struggle to represent complex lip shapes and do not model the mouth interior (see Figure \ref{fig:tracking_failure}).

Furthermore, it is desirable to introduce additional signals, such as emotion, to generate more realistic and believable video, and to offer users a level of control over the outputs. A few prior works attempt this \cite{MEAD, AudioDrivenEVP, NSPVD, EAMM, paraperas2022ned}. These methods usually either consider emotion as discrete categories \cite{MEAD, NSPVD}, which gives semantic control but lacks granularity, or learn latent encodings of emotion \cite{EAMM, AudioDrivenEVP, paraperas2022ned} which allow for fine-grained control but is not semantic and requires selecting emotions from other sources (video or audio). 

In this paper, we introduce READ Avatars, a method for generating talking head video with direct and granular control over emotion, while achieving high levels of lip sync, emotional clarity, and visual quality. We build upon 3D-based approaches, using a morphable model \cite{FLAME:SiggraphAsia2017} as an intermediate representation of the face and deferred neural rendering \cite{thies2020nvp} to achieve high visual quality. To address the above issues causing poor lip sync in 3D models, we propose two novel components. First, we add an adversarial loss to the audio-to-expression generator to alleviate the many-to-many mapping issue. Second, we overcome the challenge of representing complex lip shapes and mouth interiors with a morphable model by conditioning a neural texture on audio, encoding audio features on the surface of the mesh using a resolution-independent neural texture based on a SIREN network \cite{sitzmann2019siren}.

In summary our contributions are:

\begin{itemize}
    \item{A novel neural rendering algorithm that leverages neural textures, operates directly on UV coordinates, and can be conditioned on audio, improving the mouth interior.}
    \item{The incorporation of a GAN loss into the audio-to-expression network to improve the results by solving the many-to-many issue of audio-to-expression generation.}
    \item{A new metric for determining how well an actor's emotions are captured and reconstructed.}
\end{itemize}

\section{Related Works}

\subsection{Unconditional Audio-Driven Face Models}

The task of generating lip-synced video from audio alone, or else from audio and a reference video, known as unconditional audio-driven video generation, has been widely studied and has numerous practical applications, such as dubbing and digital avatars. There are two broad categories of unconditional models: those that use 3D priors and those 2D models that do not.

\textbf{2D Models:} Many approaches to synthesizing talking head videos from audio operate directly in the image or video domain \cite{Sun2022, Wav2Lip, ATVG}. These methods typically employ an encoder-decoder architecture. ATVG \cite{ATVG} uses audio to control 2D landmarks, which are then used to generate video with attention to highlight the parts that need editing. Wav2Lip \cite{Wav2Lip} significantly improves lip sync accuracy by minimizing the distance between the audio and generated video according to a pre-trained lip sync detection network. While the lip sync is excellent, the visual quality is poor. Recently, a context-aware transformer \cite{Sun2022} was applied to this problem, with an audio-injected refinement network that significantly improves the visual quality. However, all 2D based models to date suffer from limited visual quality. In contrast, our 3D-based method produces much higher quality videos.

\textbf{3D Guided Face Models: } Using explicit 3D supervision, ultra high-quality face models driven by various signals have been created \cite{Face2Face, DVP, NSPVD, thies2020nvp, LiveSpeechPortraits, wen2020photorealistic, LipSync3D}. These methods simplify facial synthesis by modeling the underlying 3D scene with a small set of parameters, such as a 3DMM \cite{Blanz99, 3DMM2020}, that can be directly controlled. Despite their high visual quality, these models often lack expressiveness due to the many-to-many mapping problem and the limited lip expressions of the underlying geometry.

Puppetry methods \cite{NSPVD, DVP}, and motion models \cite{EAMM, Siarohin_2019_NeurIPS} are able to somewhat solve the many-to-many issue by using a source actor to drive the expressions. This provides a signal, the source actor's expressions, which is much closer to one-to-one with the target actor. However, it is often undesirable to require a source actor to be filmed, and the resulting video processed every time the model is used. Actor-free methods such as ours are significantly more scalable. Implicit models \cite{mildenhall2020nerf, nha, IMAvatar, guo2021adnerf} augment geometry using MLP offsets, allowing for more expressive lip shapes but do not solve the many-to-many problem. 

Concurrent work \cite{Tang2022} addresses both the many-to-many issue, and the mouth interior using memory networks. However, they rely on reusing explicit pixels from the mouth region, which leads to jitter in the final videos. 

\subsection{Audio-Driven Face Models with Emotional Control}

\begin{figure*}[!ht]
    \centering
    \includegraphics[width=\textwidth]{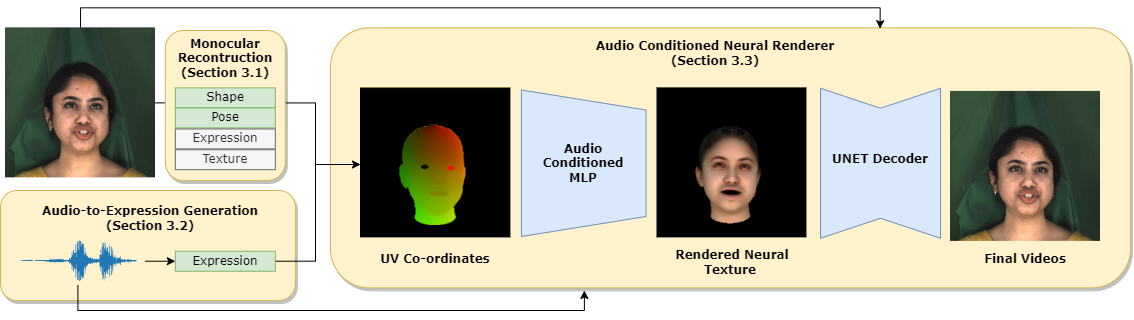}
    \caption{The READ Avatars pipeline. We train 3 separate networks. The first converts audio into expressions (Section \ref{section:a2e}). After rasterization, the second, MLP-based network converts the uv coordinates to neural texture features, conditioning on audio (Section \ref{section:NR}). The final, UNET-based network takes this rasterized neural texture, and the real video frame, and outputs lip synchronized and emotional video frames.}
    \label{fig:pipeline}
\end{figure*}

Only a small number of works have attempted to develop models that allow for explicit control of stylistic attributes, such as emotion, in generated talking head videos.

MEAD \cite{MEAD} introduces an audio-driven model with control over emotion. They trained a network to map audio to landmarks and another to convert input images to the desired emotion, and then used a UNET-based network to combine the upper face with the desired emotion and the generated landmarks to produce the final image. This method can control emotion and intensity, but lacks temporal coherence due to its frame-by-frame nature and has suboptimal lip sync. EVP \cite{AudioDrivenEVP} improves upon MEAD by adding an emotion disentanglement network to separate content and emotion in the audio and using a face-synthesis network based on vid2vid \cite{vid2vid} to produce higher quality videos with temporal consistency. However, the lip sync and emotional clarity were not always satisfactory. MEAD and EVP are the most similar to our work. These landmark based models are unable to produce as high quality results as 3D models, they also suffer from unnatural motion without the strong priors of a 3DMM.

Kim et. al. \cite{NSPVD} propose a 3D based model with control over style capable of producing highly realistic videos. The model uses a style translation network to convert the animation style of a source actor to that of a target. This, however, requires training a neural renderer for every emotion and a style translation network for every pair of emotions, which quickly becomes intractable for many styles. Similarly EAMM \cite{EAMM} uses a source actor to generate the emotional style, and a motion model to produce the lip motion. This method is unable to model the emotion in the mouth region, and as the emotional style is directly copied from a source actor, the result can appear unnatural on the target actor. NED \cite{paraperas2022ned} also manipulates the emotion of a source actor effectively using an emotional manipulation network in the parameter space of a 3DMM. Their method can semantically control the emotion, but also suffers from some unnatural motion owing to a mismatch of emotional style due to the generality of the model. 

\section{Method}

Our method consists of three stages: fitting a 3D Morphable Model (3DMM) to the input videos (Section \ref{section:track}), generating morphable model parameters from audio using adversarial training (Section \ref{section:a2e}), and training an audio-conditioned deferred neural renderer to produce the final, photo realistic video outputs (Section \ref{section:NR}). These steps are shown in Figure \ref{fig:pipeline}.

In the first stage, we fit a 3D Morphable Model \cite{FLAME:SiggraphAsia2017, Blanz99, blendshapeReview, MonocularSurvey} to the input videos using an extension of the Face2Face monocular reconstruction algorithm \cite{Face2Face}, with the modification of including blink blendshapes for both eyes. We use the implementation provided in Neural Head Avatars \cite{nha}. In the second stage, we train a neural network inspired by Pix2Pix \cite{pix2pix} to generate morphable model parameters from audio using adversarial training. This allows us to generate realistic animation sequences, even in areas that are not well correlated with audio. At this stage, we introduce a fine-grained emotional label. In the final stage, we train an audio-conditioned deferred neural renderer \cite{thies2019deferred}. Our model uses a SIREN MLP \cite{sitzmann2019siren} to directly map uv coordinates to texture features, replacing the learned neural texture of previous work \cite{thies2019deferred}, and making the task of audio conditioning much easier.

\subsection{Monocular Reconstruction}
\label{section:track}

In the first stage of our method, we aim to find a low-dimensional set of parameters that can model a video sequence $V = (V_0, \ldots V_n)$. For this purpose, we use the FLAME model \cite{FLAME:SiggraphAsia2017}, which represents explicit 3D geometry using a combination of skinned joints and blendshapes. The FLAME model can be represented as a function $\mathcal{V}$ that maps a set of parameters for shape $\beta$, expression $\psi$, and joint rotations $\theta$ onto 5023 3D vertices:

\begin{equation}
    \mathcal{V}: \mathbb{R}^{ |\beta| \times |\theta| \times |\psi|} \rightarrow \mathbb{R}^{5023 \times 3}
\end{equation}

A similar function is used to map a set of texture parameters $\alpha$ onto UV-based 2D textures.

We model the rendering process using a full perspective camera that projects a mesh $M$ onto the image plane according to:

\begin{equation}
    \hat I = \Pi(M, K, \mathbf{R}, \mathbf{t})
\end{equation}

Where $\mathbf{R}$ and $\mathbf{t}$ are the rotation and translation of the mesh in world space, and $K$ is the camera intrinsic matrix. We also model the lighting as distant and diffuse, using 3-band spherical harmonics with parameters $\gamma$. If we define $\mathbf{\pi} = (\alpha, \beta, \theta, \psi, \gamma, \mathbf{R}, \mathbf{t}, K)$ as the set of all parameters, then our objective is to find the optimal $\mathbf{\tilde \pi}$ that best fits a given image.

To fit the FLAME model to our videos, we adopt the tracking model of Neural Head Avatars \cite{nha}, which is based on Face2Face \cite{Face2Face}. This model uses differentiable rendering in Pytorch3D \cite{ravi2020pytorch3d} to minimize the $L_1$ distance between real and rendered frames, with statistical regularization over the parameters.

We assume that shape, texture, and lighting are fixed for a given actor, as our data is captured under controlled conditions. Therefore, we can first estimate $\mathbf{\pi}_{fix} = (\alpha, \gamma, K)$, the parameters that are fixed across all videos for a given subject. These parameters are then fixed for all frames in all videos of the same subject. We can then estimate $\mathbf{\pi}_{var} = (\theta, \psi, \mathbf{R}, \mathbf{t})$ on a per-frame basis. These parameters are then the target of the audio-to-parameter generator.

\subsection{Audio-to-Parameter Generator}
\label{section:a2e}

The goal of our method is to animate photo-realistic avatars using audio as the control signal. Previous approaches based on 3D Morphable Models \cite{NSPVD, thies2020nvp, wen2020photorealistic, LipSync3D} use audio-to-parameter generators to puppeteer a target actor using audio or a source video. These methods rely on neural networks with regression losses to generate a subset of the parameters. Such methods, however, suffer from the many-to-many issue when mapping audio to expressions, as multiple, equally-valid expressions can come from the same audio. Regression based losses mean that weakly correlated parameters such as upper face motion is almost entirely averaged out, while even highly correlated parameters such as the lip and jaw are over-smoothed. To address these issues, we propose an audio-to-parameter generator based on a conditional GAN \cite{goodfellow2014generative}. Our model is similar to Pix2Pix \cite{pix2pix}, using a combination of $L_1$ loss for low frequency parts of the data and an adversarial loss for increased realism. 

The input of this network is a section of audio represented as MFCC coefficients, $\mathbf{A}$, together with an explicit emotion label. The window size of the MFCC is selected to be a multiple of the video frame rate. We use an explicit emotion labeling system to introduce emotion into the generated parameters. For $N$ emotions, we use an $N-1$ dimensional label, $C$, with neutral emotion represented as a zero vector (absence of emotion). Each other emotion is assigned to a dimension and scaled by intensity, with the maximum intensity being 1. This continuous label allows for fine-grained control over the emotion. We distribute the label over the time dimension to obtain $\mathbf{C} = (C_0, \ldots, C_n)$. This label is concatenated with the MFCC audio features and serves as the input to the audio to expression generator $\mathcal{G}_a$, which produces the target parameters for each frame $(\pi_0, \ldots \pi_n)$.

\begin{equation}
(\pi_0, \ldots \pi_n) = \mathcal{G}_{A}(\mathbf{A}, \mathbf{C})
\end{equation}

The discriminator is conditional, and takes either the real or generated parameters, together with the audio and emotional label and predicts if the given parameters are real or generated.

Both the generator and discriminator networks use an encoder-temporal-decoder model, projecting the audio features into a high-dimensional latent space via a fully connected layer followed by an LSTM \cite{hochreiter1997long} and a fully connected decoder to map from this latent space to the parameters. We optimize the objective:

\begin{equation}
\label{eq:a2e}
\mathcal{L} = \mathcal{L}_1 + \lambda_{\text{GAN}}\mathcal{L}_{\text{GAN}} + \lambda_{\text{vel}}\mathcal{L}_{\text{vel}}
\end{equation}

where $\mathcal{L}_1$ is the $\ell_1$ distance between the real and predicted parameters, $\mathcal{L}_{\text{GAN}}$ is the adversarial loss and $\mathcal{L}_{\text{vel}}$ is an $\ell_1$ distance between the velocities of the output animation. The velocity loss is known to improve the temporal consistency of speech-driven animation \cite{VOCA2019}. Each $\lambda$ is a relative weight, we use $\lambda_{\text{GAN}} = 0.02$ and $\lambda_{\text{vel}} = 100$.

\subsection{Audio-Conditioned Neural Renderer}
\label{section:NR}

\begin{figure}
    \centering
    \includegraphics[width = 0.5\textwidth]{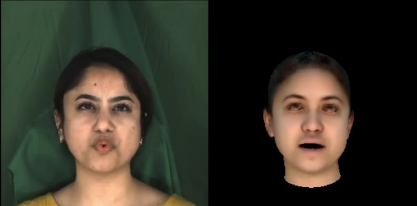}
    \caption{An example of a typical failure in the monocular reconstruction method. \textbf{\textit{(Left):}} The input frame, \textbf{\textit{(right):}} the reconstruction. Note how FLAME model lacks the expressiveness to capture certain mouth shapes, in this case an "O".}
    \label{fig:tracking_failure}
\end{figure}

\begin{table*}[!t]
    \centering
    \caption{\textbf{Quantitative comparisons to state-of-the-art}. We compare visual quality using FID, lip sync with LSE-D/C \cite{Wav2Lip} and emotional reconstruction with our metrics A/V-EMD. We compare our results with ATVG \cite{ATVG}, MEAD \cite{MEAD} and Audio-driven Emotional Video Portraits \cite{AudioDrivenEVP} (EVP)}
    \setlength{\tabcolsep}{1.2em} 
    {\renewcommand{\arraystretch}{1.2}%
    \begin{tabular}{c c c c c c}
    \hline
        \textbf{Method} & \textbf{LSE-C} $\uparrow$ & \textbf{LSE-D} $\downarrow$ & \textbf{FID} $\downarrow$ & \textbf{A-EMD} $\downarrow$ & \textbf{V-EMD} $\downarrow$ \\
    \hline
        ATVG \cite{ATVG} & $\mathbf{5.705}$ & $\mathbf{8.731}$ & $120.040$ & $0.160$ & $0.239$ \\
        MEAD \cite{MEAD} & $4.080$ & $10.569$ & $38.015$ & $0.974$ & $0.113$ \\
        EVP \cite{AudioDrivenEVP} & $4.061$ & $11.514$ & $43.972$ & $0.119$  & $0.126$ \\
        Ours & $4.431$ & $10.157$ & $\mathbf{13.600}$ & $\mathbf{0.0686}$ & $\mathbf{0.093}$ \\
        \hline
    \end{tabular}
    }
    \label{tab:compare}
\end{table*}

We next consider how to invert the parametric model fitting and produce photo-realistic video. Given a set of parameters $(\pi_i)$ corresponding to a video $\mathbf{V}$, we aim to reproduce the video as faithfully as possible. We build upon the idea of neural textures \cite{thies2019deferred}, jointly optimizing an image-to-image deferred neural renderer, and a neural texture defined in UV space. However, we find that the FLAME model is not expressive enough to represent complex lip motions (see Figure \ref{fig:tracking_failure}), and that neural textures alone are not sufficient to compensate for this. Furthermore, the FLAME mesh provides no information about the interior of the mouth including the tongue and teeth.

To address this issue, we propose audio-conditioned neural textures. The aim is to encode audio information on the surface of the mesh to allow for more complex lip shapes and mouth interiors to be learned. We replace the static, learned neural texture with a SIREN MLP \cite{sitzmann2019siren} texture network $\mathcal{T}$, which maps a uv coordinate of the rasterized meshes directly to a feature vector. This bypasses the need for texture lookup which is slow and limits the resolution of the neural textures. The use of a network also allows us to easily condition on audio by simply concatenating audio features to the uv coordinates.

We do not want the audio in the network to be biased by emotion or identity, as this will prevent us easily changing the emotion of the generated videos. To remove such information, we use the output of a Wav2Vec2 network \cite{wav2vec220, xu2021simple} pretrained to predict phoneme probabilities at a $50$ fps. The $50$ most common phonemes comprise over $99\%$ of the audio data, therefore we restrict the features to these only. Next, we resample these probabilities to $60$ fps, twice the frame rate of the video. We take a window of $W$ frames centered at the target video frame and use a small neural network $\mathcal{A}: \mathbb{R}^{2W \times 50}$ to encode the audio over this window into a single vector $\mathbf{a}_{\text{enc}} \in \mathbb{R}^{N_a}$, where $N_a$ is the dimension of the encoded audio and is a hyperparameter. The encoded audio vector is then used to condition the neural texture.

The audio encoder consists of several fully connected layers, followed by temporal convolutions, a reshaping layer that removes the time dimension, and a final fully connected decoder. The now encoded audio vector $\mathbf{a}_{enc}$ is concatenated with the UV coordinates obtained during rasterization, which serves as the input to our texture network.

The output of the texture network is a multi-channel image, which appears as a rasterization of the mesh with a neural texture. This texture encodes audio information on the surface. Similar to \cite{thies2019deferred}, we use a 16-channel neural texture, enabling representation of higher-order lighting effects. This rasterized image is then passed through a UNET \cite{UNET} based deferred neural renderer $\mathcal{R}$, which produces a photorealistic final frame, leveraging the audio features encoded on the mesh.

To address the issue of jitter in the final video, we include additional renderings for the frames in a window of length $W_{R}$ centered on the target frame. These additional renderings are based on the rasterized UV coordinates of the parameters from the frames on either side of the target frame. This results in an input with $16 + 2 {W_{R}}$ channels for the UNET decoder, which is able to smooth out the jitter over a window of frames. We have found this approach to be effective in reducing jitter in the final video.

\begin{figure}[!t]
    \centering
    \includegraphics[width=0.5\textwidth]{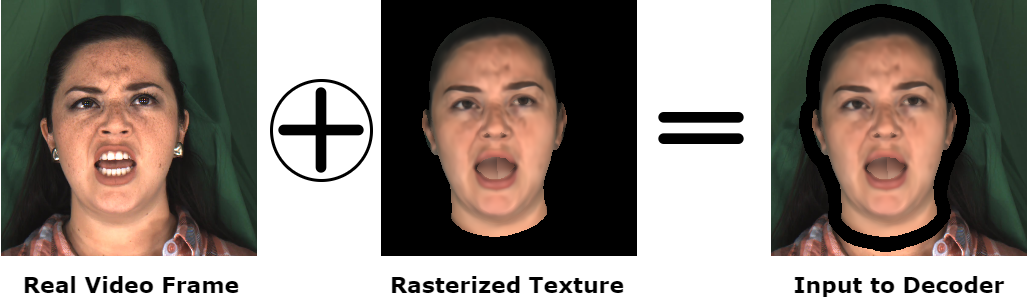}
    \caption{The input to the decoder network consists of a rendered neural texture and a real frame with a border of black pixels. Note that the texture cannot represent the mouth interior, but it is generated by the decoder.}
    \label{fig:input}
\end{figure}

In order to produce realistic and temporally consistent background in our videos, we blend the output of our texture network $\hat V_t$ with the original video frames $V_t$. We do this by using the alpha channel from the rendered mesh as a mask to separate the foreground and background. The foreground mask, $\alpha$ is expanded by a fixed number of pixels to obtain $\alpha_{\text{exp}}$. This expanded mask is zeroed out of the real frame and the foreground mask $\alpha$ is used to fill in these pixels with the rendered mesh. This process gives an input consisting of the rendered mesh in the real frame, with a border that the decoder can inpaint. This is best shown in Figure \ref{fig:input}.

\begin{equation}
 V^*_t = \alpha \hat V_t + (1 - \alpha_{\text{exp}}) V_t
\end{equation}

To train the audio encoder, texture network, and deferred neural renderer, we optimize the following objective function end-to-end:

\begin{equation}
    \mathcal{L}(V^*, V) = \lambda_{1}\mathcal{L}_1 + \lambda_{\text{VGG}}\mathcal{L}_{\text{VGG}} + \lambda_{\text{GAN}}\mathcal{L}_{\text{GAN}}
\end{equation}

 Here, $\mathcal{L}_1$ is the $\ell_1$ distance between the real and generated frames, $\mathcal{L}_{VGG}$ is a VGG-based style loss \cite{johnson2016perceptual}, and $\mathcal{L}_{\text{GAN}}$ is an adversarial loss. The hyperparameters $\lambda_1$, $\lambda_{\text{VGG}}$, and $\lambda_{\text{GAN}}$ are used to weight the importance of each loss. We use $\lambda_1 = \lambda_{\text{VGG}} = 1$ and $\lambda_{\text{GAN}} = 0.01$.

\subsection{Implementation Details}

\begin{figure*}[!t]
    \centering
    \includegraphics[width=\textwidth]{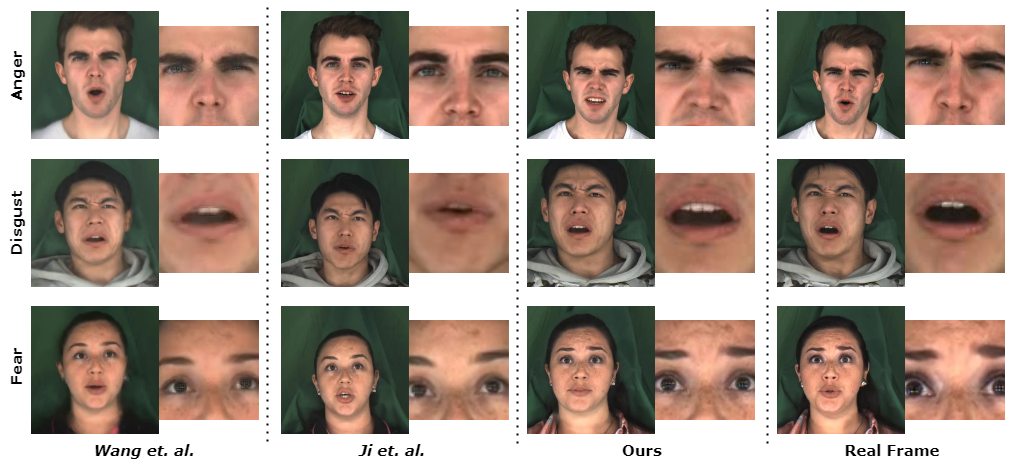}
    \caption{We compare our results to those of MEAD \cite{MEAD} and EVP \cite{AudioDrivenEVP}. Our results are of much higher visual quality that those of MEAD, the zoomed in regions demonstrate that our method produces more convincing and accurate emotions compared to EVP.}
    \label{fig:comapre}
\end{figure*}

To prepare the data for our method, we first crop every frame to a square shape of $256$ pixels. We do this by estimating a bounding box for each frame with padding, then finding the smallest square that covers the union of these boxes. We reshape this square to the desired resolution. We implement our pipeline in Pytorch and Pytorch3D. Our models are trained on a single NVIDIA RTX3080 graphics card. All networks are optimized using Adam \cite{Adam} with a learning rate of $0.0001$. The renderer is trained for 5 epochs, taking about 15 hours, while the audio-to-expression generator is trained for 10 epochs taking around 5 hours. We use the LSGAN formulation for all adversarial training \cite{mao2017least}.

\section{Results}

\subsection{Dataset}

\begin{figure*}[h]
    \centering
    \includegraphics[width=0.9\textwidth]{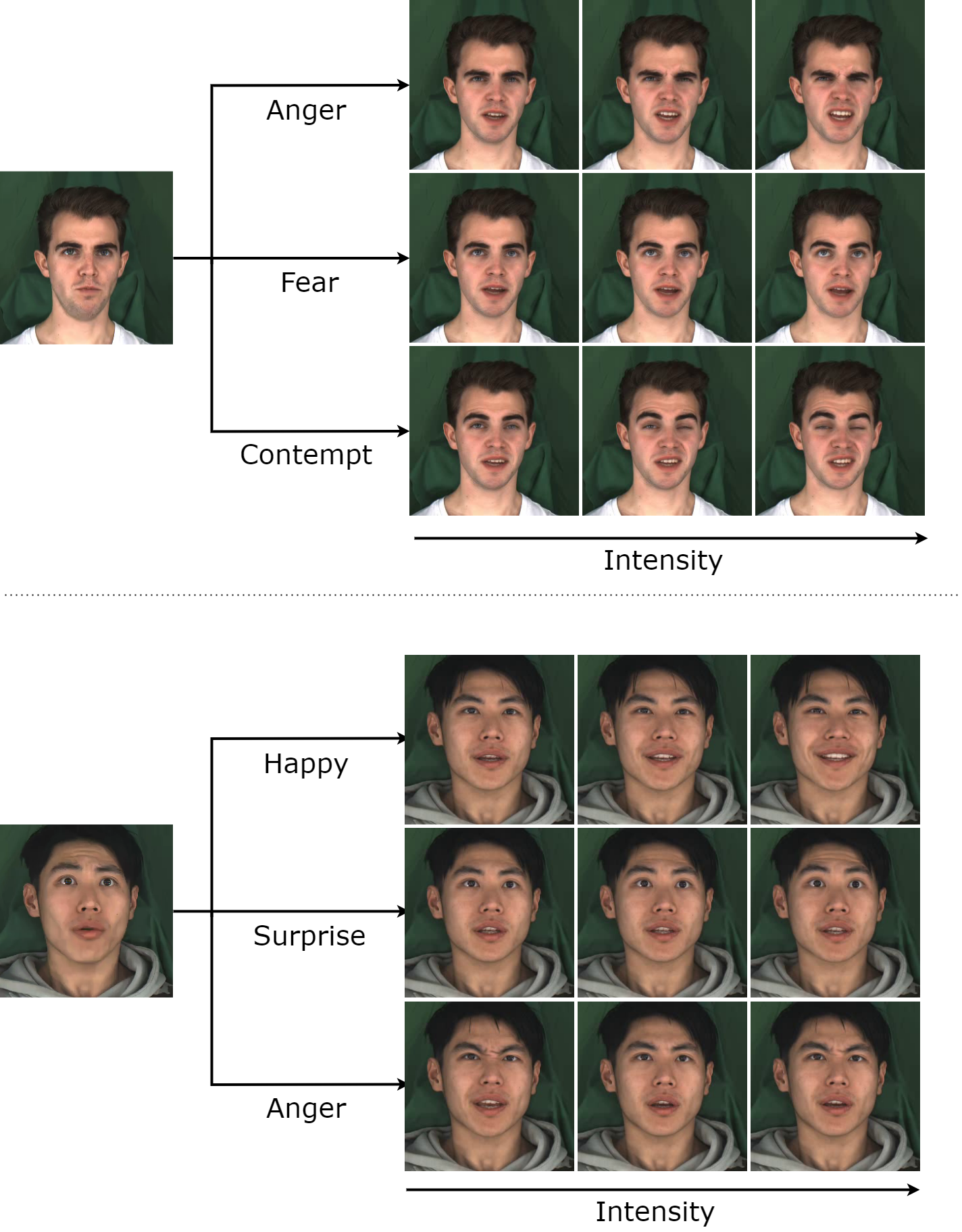}
    \caption{Our method allows for fine-grained control on multiple subjects. Here we show two subjects in with three emotions and three levels of intensity.}
    \label{fig:extra_title}
\end{figure*}

We use the MEAD dataset \cite{MEAD} for our experiments, which includes 60 actors (both male and female) speaking 30 sentences in 8 different emotions at 3 levels of intensity, recorded from multiple angles. For this work, we only use the front-facing camera footage. We follow the train-test split outlined in MEAD and train models for 4 of these subjects. Figure \ref{fig:extra_title} shows a selection of results coming from our method, across multiple emotions and intensities.

\subsection{Quantitative metrics}

In order to evaluate the performance of our method, we consider three qualities: visual quality, lip sync, and emotional clarity.

For visual quality, we use the Fréchet Inception Distance (FID) metric to measure the similarity of the generated frames to the ground truth. We crop the frames tightly around the face region in order to avoid biasing the results towards our method, which uses the ground truth background. To measure lip synchronization, we use the Lip Sync Error (LSE) metrics introduced in wav2lip \cite{Wav2Lip}. These metrics are calculated using a pre-trained syncnet and include LSE-D, which measures the minimum distance between audio and video features, and LSE-C, which measures the confidence that the audio and video are synchronized. To measure emotional clarity, it is not enough to measure differences at the frame level, as the intensity of emotion naturally varies over a video. We therefore, introduce a new metric for emotional clarity that measures the differences between distributions of emotion. This metric is based on a pre-trained EmoNet model \cite{emonet2021estimation}. We predict the valence and arousal for each frame, and compare the distributions of these values between the generated and ground truth videos. We approximate the distance between these distributions using the Earth Movers Distance, and compute this distance for each subject and emotion separately, taking the average to obtain the valence Emotional Mean Distance (V-EMD) and arousal Emotional Mean Distance (A-EMD) metrics.  

\subsection{Comparisons to State-of-the-Art}

\begin{table*}[!t]
    \caption{Results of the user study. We ask users to select their preference between our video and each of the competitors for four criteria. Where ours is preferred strongly (weakly) we denote the result $++$ ($+$), where there is no preference, 0 and where the other method is preferred strongly (weakly), $--$ ($-$). The data in all but the rightmost column is in percentages. Note the data is rounded and may not add to 100\%.}
    \label{tab:user_study}
    \centering
    \begin{tabular}{c c c c c c c}
    \hline
        Statement &  $--$ & $-$ & 0 & $+$ & $++$ & mean \\
     \hline
         Ours $>$ MEAD (lip-sync ) & $1$ & $12$ & $14$ & $28$ & $44$ & $+1.02$ \\
         Ours $>$ MEAD (visual quality) & $0$ & $3$ & $9$ & $23$ & $65$ & $+1.49$ \\
         Ours $>$ MEAD (naturalness) & $1$ & $1$ & $15$ & $23$ & $65$ & $+1.39$ \\
         Ours $>$ MEAD (emotion) & $1$ & $3$ & $21$ & $36$ & $38$ & $+1.06$ \\
     \hline
        Ours $>$ EVP (lip-sync ) & $7$ & $16$ & $22$ & $42$ & $11$ & $+0.50$ \\
         Ours $>$ EVP (visual quality) & $2$ & $22$ & $23$ & $42$ & $11$ & $+0.39$ \\
         Ours $>$ EVP (naturalness) & $7$ & $17$ & $18$ & $26$ & $38$ & $+0.49$ \\
         Ours $>$ EVP (emotion) & $4$ & $8$ & $22$ & $26$ & $38$ & $+0.87$ \\
     \hline
        Ours $>$ Real (lip-sync ) & $69$ & $28$ & $3$ & $1$ & $0$ & $-1.64$ \\
         Ours $>$ Real (visual quality) & $33$ & $41$ & $23$ & $2$ & $0$ & $-1.05$ \\
         Ours $>$ Real (naturalness) & $53$ & $38$ & $8$ & $1$ & $0$ & $-1.43$ \\
         Ours $>$ Real (emotion) & $40$ & $38$ & $19$ & $3$ & $1$ & $-1.13$ \\
     \hline
    \end{tabular}
\end{table*}

We compare our method to several state-of-the-art audio-driven avatar models. Our main comparisons are with Audio-Driven Expressive Video Generation (EVP) \cite{AudioDrivenEVP} and Multimodal Emotion-aware Dataset (MEAD) \cite{MEAD}. MEAD uses an audio-to-landmark LSTM, an emotion transformer to alter the audio-driven landmarks to any given emotion, and a final UNET-based model to produce output frames from the emotional landmarks. AudioDrivenEVP improves on this approach by designing a disentanglement model to separate audio into emotion and content, which is then used with a landmark alignment method to control for pose, and a video-to-video network that produces high-quality and temporally stable video from landmarks. We also compare to ATVG \cite{ATVG}, a 2D-based method that excels in lip synchronization but has poor visual quality, and is unable to edit emotion.

\textbf{Quantitative:} The results of these comparisons are shown in Table \ref{tab:compare}. Our results outperform all competitors on visual quality (FID) and emotional reconstruction (A/V-EMD). While our method is not able to reach the lip-sync quality of the unconditional ATVG \cite{ATVG}, it has far better visual quality and emotional clarity. We outperform both methods capable of controlling emotion: MEAD \cite{MEAD} and EVP \cite{AudioDrivenEVP} on lip sync. 

\begin{table*}[!t]
    \centering
    \caption{\textbf{Ablation study}. We compare visual quality using FID, lip sync with LSE-D/C \cite{Wav2Lip} and emotional reconstruction with our metrics A/V-EMD. We compare our full model to the same model both without the GAN loss in the audio-to-expression generator and without the audio conditioned neural texture.}
    \setlength{\tabcolsep}{1.2em} 
    {\renewcommand{\arraystretch}{1.2}%
    \begin{tabular}{c c c c c c}
    \hline
        \textbf{Method} & \textbf{LSE-C} $\uparrow$ & \textbf{LSE-D} $\downarrow$ & \textbf{FID} $\downarrow$ & \textbf{A-EMD} $\downarrow$ & \textbf{V-EMD} $\downarrow$ \\
    \hline
        Ours & $\mathbf{4.431}$ & $\mathbf{10.157}$ & $13.600$ & $\mathbf{0.069}$ & $0.093$ \\
        Ours w/o GAN loss & $4.047$ & $10.446$ & $\mathbf{12.587}$ & $0.079$ & $\mathbf{0.090}$ \\
        Ours w/o audio texture & $4.175$ & $10.398$ & $15.96$ & $0.069$ & $0.96$ \\
        \hline
    \end{tabular}
    }
    \label{tab:ablation}
\end{table*}

\textbf{Qualitative:} Figure \ref{fig:comapre} shows our results in comparison to MEAD \cite{MEAD} and EVP \cite{AudioDrivenEVP}. Our results show clearly better visual quality than MEAD. Compared with EVP our method is capable of producing emotion that is much clearer and more closely matches the real videos, the expanded regions highlight this. In particular, it can be seen that the eyebrows convey the target emotion far better in our method. Additional results can be found in the supplementary video that further demonstrate the advantages of our method.  

\textbf{User Study:} To gauge the subjective quality of our generated avatars, we conducted a user study. We selected four subjects and generated five videos in each of the eight emotions for a total of 40 videos. The background and pose parameters for these videos were taken from the longest video of the target emotion in the training set. We conducted our user study to compare our work to that of MEAD \cite{MEAD}, EVP \cite{AudioDrivenEVP} and the real videos. We perform a two alternative forced choice study, pairing each of our videos with its counterpart from the alternatives. The users are shown both of the videos in a random order, together with the target emotion. We ask users to select which of the two videos is better in four categories: lip sync, visual quality, naturalness and emotional clarity. Users are able to specify if they prefer our video strongly $++$, weakly $+$, the other video strongly $--$ or weakly $-$ or if they find them equal, 0. A total of $10$ users completed the study. The results are shown in Table \ref{tab:user_study}. Our method strongly outperforms MEAD across all categories. Compared with EVP, our method is also preferred across all categories. However, this preference is weaker for lip sync and natural quality, but much stronger for emotional clarity. The user study shows that our work still does not reach the quality of real video, suggesting there is still room for future work.

\subsection{Ablation Study}

\begin{figure}[!t]
    \centering
    \includegraphics[width=0.5\textwidth]{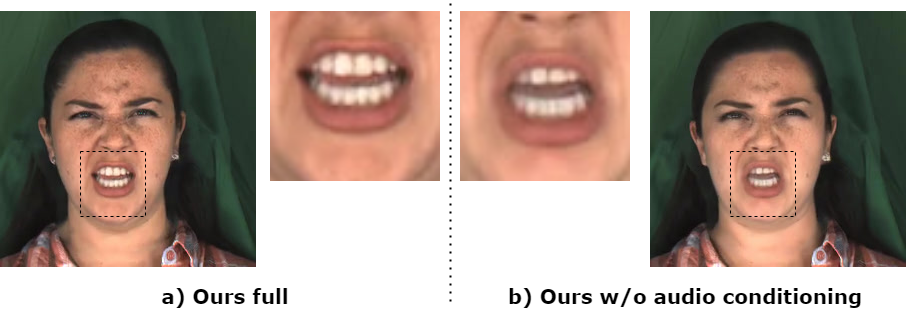}
    \caption{The addition of audio conditioning in the neural texture improves the quality of the resulting frame, particularly in the mouth interior.}
    \label{fig:ablation_NT}
\end{figure}

We also perform an ablation study for both the adversarial loss and the audio conditioned neural texture. The results of this comparison are shown in Table \ref{tab:ablation}. The inclusion the adversarial loss improves the lip-sync at the cost of a small loss in visual quality. We note that the adversarial loss has little effect on the valence metric, but a stronger effect on the arousal. We hypothesize that this difference is due to the fluctuation in the intensity of emotion being smoothed with a pure regression loss. For the audio-conditioned neural texture, we compare our work to a static, neural texture \cite{thies2019deferred}. Our method improves both the visual quality and lip-sync, with small improvements in the emotional reconstruction. As expected, the improvements of our audio conditioning are most notable in the mouth interior. This is because the audio allows the decoder to disambiguate the multiple mouth interiors that could be represented by the same underlying morphable model geometry. Figure \ref{fig:ablation_NT} shows this improvement.

\section{Conclusion}

We present a new method for producing audio driven avatars with control over emotion. We have used a 3D-based pipeline with the addition of an adversarial loss in the audio-to-expression generator and an audio-conditioned, resolution independent neural texture. Our method alleviates the many-to-many problem in conditioned, audio-driven video generation and surpasses state-of-the-art for lip sync and visual quality, as well as emotional reconstruction, as highlighted by our novel metric. Our comprehensive solution can be used for diverse applications.

\begin{figure}
    \centering
    \includegraphics[width=0.5\textwidth]{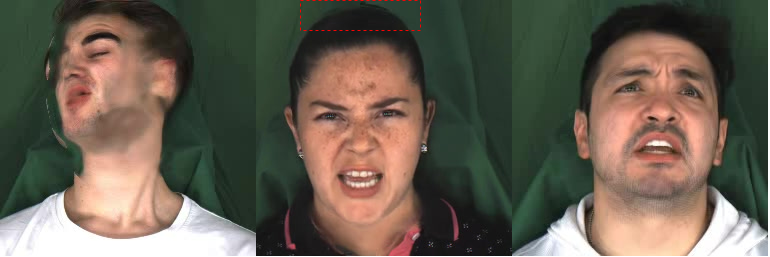}
    \caption{Failure cases of our method. \textbf{Left \& middle:} When the pose of the target video is significantly different from the training data, artifacts occur. \textbf{Right:} When the tracking is inaccurate, our model produces blurry results.}
    \label{fig:failure}
\end{figure}

\textbf{Limitations: } Our model sometimes suffers when using extreme poses (Figure \ref{fig:failure}). Furthermore, as we use reference videos to control for pose and background, length of the generated videos is limited. Future work will look to address arbitrary length video generation, potentially by considering pose generation. It is also worth investigating other models that address many-to-many generation, such as diffusion models \cite{ho2020denoising, Yang2022DiffusionMA}.

\textbf{Ethical Implications: }The ability to create synthetic digital humans comes with a serious potential for misuse. In particular, works such as ours could be altered in order to produce convincing misinformation. For this reason, we do not make our pipeline available to the general public. However, we are willing to share code with other researchers.

\section*{Acknowledgments}

This work was supported in part by UKRI grant reference EP/S023437/1. 

\newpage

\bibliographystyle{plain}
\bibliography{main}

\begin{thebibliography}{10}

\bibitem{wav2vec220}
Alexei Baevski, Yuhao Zhou, Abdelrahman Mohamed, and Michael Auli.
\newblock wav2vec 2.0: A framework for self-supervised learning of speech
  representations.
\newblock {\em Advances in Neural Information Processing Systems},
  33:12449--12460, 2020.

\bibitem{Blanz99}
Volker Blanz and Thomas Vetter.
\newblock A morphable model for the synthesis of 3d faces.
\newblock In {\em Proceedings of the 26th Annual Conference on Computer
  Graphics and Interactive Techniques}, SIGGRAPH '99, page 187–194, USA,
  1999. ACM Press/Addison-Wesley Publishing Co.

\bibitem{ATVG}
Lele Chen, Ross~K. Maddox, Zhiyao Duan, and Chenliang Xu.
\newblock Hierarchical cross-modal talking face generation with dynamic
  pixel-wise loss.
\newblock In {\em Proceedings of the IEEE/CVF Conference on Computer Vision and
  Pattern Recognition (CVPR)}, June 2019.

\bibitem{VOCA2019}
Daniel Cudeiro, Timo Bolkart, Cassidy Laidlaw, Anurag Ranjan, and Michael
  Black.
\newblock Capture, learning, and synthesis of {3D} speaking styles.
\newblock In {\em Proceedings IEEE Conf. on Computer Vision and Pattern
  Recognition (CVPR)}, pages 10101--10111, 2019.

\bibitem{3DMM2020}
Bernhard Egger, William A.~P. Smith, Ayush Tewari, Stefanie Wuhrer, Michael
  Zollhoefer, Thabo Beeler, Florian Bernard, Timo Bolkart, Adam Kortylewski,
  Sami Romdhani, Christian Theobalt, Volker Blanz, and Thomas Vetter.
\newblock 3d morphable face models—past, present, and future.
\newblock {\em ACM Trans. Graph.}, 2020.

\bibitem{goodfellow2014generative}
Ian~J. Goodfellow, Jean Pouget-Abadie, Mehdi Mirza, Bing Xu, David
  Warde-Farley, Sherjil Ozair, Aaron Courville, and Yoshua Bengio.
\newblock Generative adversarial networks, 2014.

\bibitem{nha}
Philip-William Grassal, Malte Prinzler, Titus Leistner, Carsten Rother,
  Matthias Nie{\ss}ner, and Justus Thies.
\newblock Neural head avatars from monocular rgb videos.
\newblock {\em arXiv preprint arXiv:2112.01554}, 2021.

\bibitem{guo2021adnerf}
Yudong Guo, Keyu Chen, Sen Liang, Yongjin Liu, Hujun Bao, and Juyong Zhang.
\newblock Ad-nerf: Audio driven neural radiance fields for talking head
  synthesis.
\newblock In {\em {IEEE/CVF} International Conference on Computer Vision
  (ICCV)}, 2021.

\bibitem{ho2020denoising}
Jonathan Ho, Ajay Jain, and Pieter Abbeel.
\newblock Denoising diffusion probabilistic models.
\newblock {\em arXiv preprint arxiv:2006.11239}, 2020.

\bibitem{hochreiter1997long}
Sepp Hochreiter and J{\"u}rgen Schmidhuber.
\newblock Long short-term memory.
\newblock {\em Neural computation}, 9(8):1735--1780, 1997.

\bibitem{pix2pix}
Phillip Isola, Jun-Yan Zhu, Tinghui Zhou, and Alexei~A. Efros.
\newblock Image-to-image translation with conditional adversarial networks,
  2016.

\bibitem{EAMM}
Xinya Ji, Hang Zhou, Kaisiyuan Wang, Qianyi Wu, Wayne Wu, Feng Xu, and Xun Cao.
\newblock Eamm: One-shot emotional talking face via audio-based emotion-aware
  motion model.
\newblock In {\em ACM SIGGRAPH 2022 Conference Proceedings}, SIGGRAPH '22, New
  York, NY, USA, 2022. Association for Computing Machinery.

\bibitem{AudioDrivenEVP}
Xinya Ji, Hang Zhou, Kaisiyuan Wang, Wayne Wu, Chen~Change Loy, Xun Cao, and
  Feng Xu.
\newblock Audio-driven emotional video portraits.
\newblock In {\em Proceedings of the IEEE Conference on Computer Vision and
  Pattern Recognition (CVPR)}, 2021.

\bibitem{johnson2016perceptual}
Justin Johnson, Alexandre Alahi, and Li~Fei-Fei.
\newblock Perceptual losses for real-time style transfer and super-resolution.
\newblock In {\em European conference on computer vision}, pages 694--711.
  Springer, 2016.

\bibitem{NSPVD}
Hyeongwoo Kim, Mohamed Elgharib, Hans-Peter Zoll{\"o}fer, Michael~Seidel, Thabo
  Beeler, Christian Richardt, and Christian Theobalt.
\newblock Neural style-preserving visual dubbing.
\newblock {\em ACM Transactions on Graphics (TOG)}, 38(6):178:1--13, 2019.

\bibitem{DVP}
Hyeongwoo Kim, Pablo Garrido, Ayush Tewari, Weipeng Xu, Justus Thies, Matthias
  Nie{\ss}ner, Patrick P{\'e}rez, Christian Richardt, Michael Zoll{\"o}fer, and
  Christian Theobalt.
\newblock Deep video portraits.
\newblock {\em ACM Transactions on Graphics (TOG)}, 37(4):163, 2018.

\bibitem{Adam}
Diederik~P. Kingma and Jimmy Ba.
\newblock Adam: {A} method for stochastic optimization.
\newblock In Yoshua Bengio and Yann LeCun, editors, {\em 3rd International
  Conference on Learning Representations, {ICLR} 2015, San Diego, CA, USA, May
  7-9, 2015, Conference Track Proceedings}, 2015.

\bibitem{LipSync3D}
Avisek Lahiri, Vivek Kwatra, Christian Frueh, John Lewis, and Chris Bregler.
\newblock Lipsync3d: Data-efficient learning of personalized 3d talking faces
  from video using pose and lighting normalization.
\newblock In {\em Proceedings of the IEEE/CVF Conference on Computer Vision and
  Pattern Recognition (CVPR)}, 2021.

\bibitem{blendshapeReview}
J.~P. Lewis, Ken Anjyo, Taehyun Rhee, Mengjie Zhang, Fred Pighin, and Zhigang
  Deng.
\newblock {Practice and Theory of Blendshape Facial Models}.
\newblock In Sylvain Lefebvre and Michela Spagnuolo, editors, {\em Eurographics
  2014 - State of the Art Reports}. The Eurographics Association, 2014.

\bibitem{FLAME:SiggraphAsia2017}
Tianye Li, Timo Bolkart, Michael.~J. Black, Hao Li, and Javier Romero.
\newblock Learning a model of facial shape and expression from {4D} scans.
\newblock {\em ACM Transactions on Graphics, (Proc. SIGGRAPH Asia)},
  36(6):194:1--194:17, 2017.

\bibitem{LiveSpeechPortraits}
Yuanxun Lu, Jinxiang Chai, and Xun Cao.
\newblock Live speech portraits: Real-time photorealistic talking-head
  animation.
\newblock {\em ACM Trans. Graph.}, 40(6), dec 2021.

\bibitem{mao2017least}
Xudong Mao, Qing Li, Haoran Xie, Raymond~YK Lau, Zhen Wang, and Stephen
  Paul~Smolley.
\newblock Least squares generative adversarial networks.
\newblock In {\em Proceedings of the IEEE international conference on computer
  vision}, pages 2794--2802, 2017.

\bibitem{mildenhall2020nerf}
Ben Mildenhall, Pratul~P. Srinivasan, Matthew Tancik, Jonathan~T. Barron, Ravi
  Ramamoorthi, and Ren Ng.
\newblock Nerf: Representing scenes as neural radiance fields for view
  synthesis.
\newblock In {\em ECCV}, 2020.

\bibitem{paraperas2022ned}
Foivos Paraperas~Papantoniou, Panagiotis~P. Filntisis, Petros Maragos, and
  Anastasios Roussos.
\newblock Neural emotion director: Speech-preserving semantic control of facial
  expressions in "in-the-wild" videos.
\newblock In {\em Proceedings of the IEEE/CVF Conference on Computer Vision and
  Pattern Recognition (CVPR)}, 2022.

\bibitem{Wav2Lip}
K~R Prajwal, Rudrabha Mukhopadhyay, Vinay~P. Namboodiri, and C.V. Jawahar.
\newblock A lip sync expert is all you need for speech to lip generation in the
  wild.
\newblock In {\em Proceedings of the 28th ACM International Conference on
  Multimedia}, MM '20, page 484–492, New York, NY, USA, 2020. Association for
  Computing Machinery.

\bibitem{ravi2020pytorch3d}
Nikhila Ravi, Jeremy Reizenstein, David Novotny, Taylor Gordon, Wan-Yen Lo,
  Justin Johnson, and Georgia Gkioxari.
\newblock Accelerating 3d deep learning with pytorch3d.
\newblock {\em arXiv:2007.08501}, 2020.

\bibitem{UNET}
Olaf Ronneberger, Philipp Fischer, and Thomas Brox.
\newblock U-net: Convolutional networks for biomedical image segmentation.
\newblock In Nassir Navab, Joachim Hornegger, William~M. Wells, and
  Alejandro~F. Frangi, editors, {\em Medical Image Computing and
  Computer-Assisted Intervention -- MICCAI 2015}, pages 234--241, Cham, 2015.
  Springer International Publishing.

\bibitem{Siarohin_2019_NeurIPS}
Aliaksandr Siarohin, Stéphane Lathuilière, Sergey Tulyakov, Elisa Ricci, and
  Nicu Sebe.
\newblock First order motion model for image animation.
\newblock In {\em Conference on Neural Information Processing Systems
  (NeurIPS)}, December 2019.

\bibitem{sitzmann2019siren}
Vincent Sitzmann, Julien~N.P. Martel, Alexander~W. Bergman, David~B. Lindell,
  and Gordon Wetzstein.
\newblock Implicit neural representations with periodic activation functions.
\newblock In {\em Proc. NeurIPS}, 2020.

\bibitem{Sun2022}
Yasheng Sun, Hang Zhou, Kaisiyuan Wang, Qianyi Wu, Zhibin Hong, Jingtuo Liu,
  Errui Ding, Jingdong Wang, Ziwei Liu, and Koike Hideki.
\newblock Masked lip-sync prediction by audio-visual contextual exploitation in
  transformers.
\newblock In {\em SIGGRAPH Asia 2022 Conference Papers}, SA '22, New York, NY,
  USA, 2022. Association for Computing Machinery.

\bibitem{Tang2022}
Anni Tang, Tianyu He, Xu~Tan, Jun Ling, Runnan Li, Sheng Zhao, Li~Song, and
  Jiang Bian.
\newblock Memories are one-to-many mapping alleviators in talking face
  generation, 2022.

\bibitem{thies2020nvp}
Justus Thies, Mohamed Elgharib, Ayush Tewari, Christian Theobalt, and Matthias
  Nie{\ss}ner.
\newblock Neural voice puppetry: Audio-driven facial reenactment.
\newblock {\em ECCV 2020}, 2020.

\bibitem{thies2019deferred}
Justus Thies, Michael Zollh{\"o}fer, and Matthias Nie{\ss}ner.
\newblock Deferred neural rendering: Image synthesis using neural textures.
\newblock {\em ACM Transactions on Graphics (TOG)}, 38(4):1--12, 2019.

\bibitem{Face2Face}
Justus Thies, Michael Zollh\"{o}fer, Marc Stamminger, Christian Theobalt, and
  Matthias Nie\ss{}ner.
\newblock Face2face: Real-time face capture and reenactment of rgb videos.
\newblock {\em Commun. ACM}, 62(1):96–104, dec 2018.

\bibitem{emonet2021estimation}
Antoine Toisoul, Jean Kossaifi, Adrian Bulat, Georgios Tzimiropoulos, and Maja
  Pantic.
\newblock Estimation of continuous valence and arousal levels from faces in
  naturalistic conditions.
\newblock {\em Nature Machine Intelligence}, 2021.

\bibitem{vougioukas2018end}
Konstantinos Vougioukas, Stavros Petridis, and Maja Pantic.
\newblock End-to-end speech-driven facial animation with temporal gans.
\newblock {\em arXiv preprint arXiv:1805.09313}, 2018.

\bibitem{MEAD}
Kaisiyuan Wang, Qianyi Wu, Linsen Song, Zhuoqian Yang, Wayne Wu, Chen Qian, Ran
  He, Yu~Qiao, and Chen~Change Loy.
\newblock Mead: A large-scale audio-visual dataset for emotional talking-face
  generation.
\newblock In {\em ECCV}, 2020.

\bibitem{vid2vid}
Ting-Chun Wang, Ming-Yu Liu, Jun-Yan Zhu, Guilin Liu, Andrew Tao, Jan Kautz,
  and Bryan Catanzaro.
\newblock Video-to-video synthesis.
\newblock In {\em Advances in Neural Information Processing Systems (NeurIPS)},
  2018.

\bibitem{wen2020photorealistic}
Xin Wen, Miao Wang, Christian Richardt, Ze-Yin Chen, and Shi-Min Hu.
\newblock Photorealistic audio-driven video portraits.
\newblock {\em IEEE Transactions on Visualization and Computer Graphics},
  26(12):3457--3466, 2020.

\bibitem{xu2021simple}
Qiantong Xu, Alexei Baevski, and Michael Auli.
\newblock Simple and effective zero-shot cross-lingual phoneme recognition.
\newblock {\em arXiv preprint arXiv:2109.11680}, 2021.

\bibitem{Yang2022DiffusionMA}
Ling Yang, Zhilong Zhang, Yang Song, Shenda Hong, Runsheng Xu, Yue Zhao,
  Yingxia Shao, Wentao Zhang, Bin Cui, and Ming-Hsuan Yang.
\newblock Diffusion models: A comprehensive survey of methods and applications.
\newblock {\em arXiv preprint arXiv:2209.00796}, 2022.

\bibitem{TextBasedEdit}
Xinwei Yao, Ohad Fried, Kayvon Fatahalian, and Maneesh Agrawala.
\newblock Iterative text-based editing of talking-heads using neural
  retargeting.
\newblock {\em ACM Trans. Graph.}, 40(3), aug 2021.

\bibitem{IMAvatar}
Yufeng Zheng, Victoria~Fernández Abrevaya, Marcel~C. Bühler, Xu~Chen,
  Michael~J. Black, and Otmar Hilliges.
\newblock {I} {M} {Avatar}: Implicit morphable head avatars from videos.
\newblock In {\em Computer Vision and Pattern Recognition (CVPR)}, 2022.

\bibitem{MonocularSurvey}
M.~Zollhöfer, J.~Thies, P.~Garrido, D.~Bradley, Thabo Beeler, P.~Pérez,
  M.~Stamminger, M.~Nießner, and C.~Theobalt.
\newblock State of the art on monocular 3d face reconstruction, tracking, and
  applications.
\newblock {\em Computer Graphics Forum}, 37:523--550, 05 2018.

\end{thebibliography}

\end{document}